\documentclass[11pt,a4paper]{article}
\usepackage[hyperref]{acl2019}
\usepackage{times}
\usepackage{latexsym}
\usepackage{graphicx} 
\usepackage{url}
\usepackage{multirow}
\usepackage{amsmath}

\aclfinalcopy 
\def\aclpaperid{896} 

\title{Does It Make Sense? And Why? A Pilot Study for \\ Sense Making and Explanation}

\author{Cunxiang Wang\textsuperscript{1,4}, Shuailong Liang\textsuperscript{2}, Yue Zhang\textsuperscript{1}, Xiaonan Li\textsuperscript{3} and Tian Gao\textsuperscript{4}\\
\textsuperscript{1}School of Engineering, Westlake University, China\\
\textsuperscript{2}Singapore University of Technology and Design, Singapore\\
\textsuperscript{3}School of Computer Science and Technology, Xidian University, China\\
\textsuperscript{4}College of Computer Science and Technology, Zhejiang University, China\\
  {\tt wangcunxiang@westlake.edu.cn},
  {\tt yue.zhang@wias.org.cn}\\
  {\tt shuailong\_liang@mymail.sutd.edu.sg}\\
  {\tt lixiaonan\_xdu@outlook.com},
  {\tt gaotian@zju.edu.cn} \\
  }

\date{}

\begin{document}
\maketitle
\begin{abstract} 
  Introducing common sense to natural language understanding systems has received increasing research attention. It remains a fundamental question on how to evaluate whether a system has a sense making capability. Existing benchmarks measures commonsense knowledge indirectly and without explanation. In this paper, we release a benchmark to directly test whether a system can differentiate natural language statements that make sense from those that do not make sense. In addition, a system is asked to identify the most crucial reason why a statement does not make sense. We evaluate models trained over large-scale language modeling tasks as well as human performance, showing that there are different challenges for system sense making.
\end{abstract}

\section{Introduction}

Natural Language Understanding (NLU) has received increasing research attention in recent years. 
With language models trained on large corpora \citep{ELMO,BERT}, 
algorithms show better performance than humans on some benchmarks \citep{R-NET, BERT}.
Compared to humans, however, most end-to-end trained systems are rather weak on common sense. 
For example, it is straightforward for a human to understand that \emph{someone can put a turkey into a fridge} but \emph{he can never put an elephant into a fridge} with basic commonsense reasoning, but it can be non-trivial for a system to tell the difference. Arguably, commonsense reasoning should be a central capability in a practical NLU system \cite{logical}; it is, therefore, important to be able to evaluate how well a model can do for sense making.



Existing datasets test common sense indirectly through tasks that require extra knowledge, such as co-reference resolution \cite{WSC2012,WSC2015}, subsequent event prediction \cite{COPA,JOCI,SWAG}, or reading comprehension \cite{RocStories2016,SemEval-2018-Task-11}. They verify whether a system is equipped with common sense by testing whether it can give a correct answer where the input does not contain such knowledge. However, there are two limitations to such benchmarks. First, they do not give a direct metric to quantitatively measure sense making capability. Second, they do not explicitly identify the key factors required in a sense making process.


\begin{figure}[tb]
  \center{\includegraphics[width=8cm]  {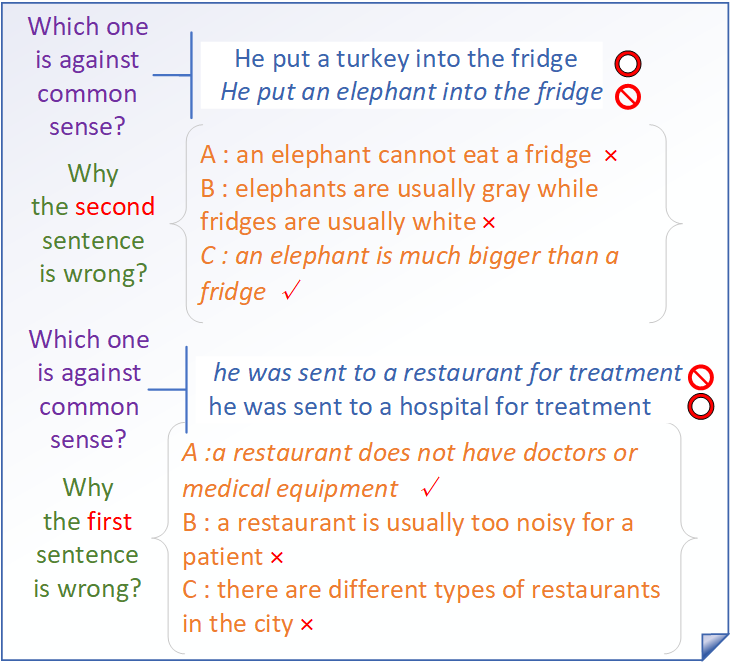}}  
  \caption{\label{1}  Samples of our dataset}
  \vspace{-3.3mm}
\end{figure}

We address these issues by creating a testset for direct sense making. As shown in Figure 1, the benchmark consists of two subtasks. The first task is to choose from two natural language statements with similar wordings which one makes sense and which one does not make sense; The second task is to find the key reason why a given statement does not make sense. Version 1.0 of our dataset, which will be released with this paper, contains 2021 instances for each subtask, manually labeled by 7 annotators. Human performance on the benchmark is 99.1\% for the Sen-Making task and 97.3\% for the Explanation task.

In the Sen-Making task, we use statement pair classification rather than labelling each statement `true' or `false' in the absolute sense because it is easy to cite a counterexample for any single `true' or `false' statement. For example, `toy elephant' for `he put an elephant into fridge'. But it is apparent that `he put a turkey into fridge' is more reasonable than `he put an elephant into fridge'.

In this pilot study, we evaluate contextualized representations trained over large-scale language modeling tasks on our benchmark. Results show that there is still a large gap behind human performance despite that the models are trained over 100 million natural language sentences. Detailed examination shows that inference remains a challenge for such systems. To our knowledge, our dataset has the most direct decision-making process in commonsense reasoning and is the first one asking reasons behind the decision making process.

Note that there has been dataset which focus on non-linguistic world knowledge plausibility \citep{wang2018modeling} or only limited attributes or actions of physical knowledge like verbphysics \cite{verbphysics}. They are related to our dataset but serve robotic research mainly. Our dataset is the first benchmark for direct linguistic sense making and explanation. We hope this benchmark can promote commonsense reasoning by the NLP community, and further applied on other applications such as machine translation and dialogue. Besides, we also expect that this work could be instructive on enhancing interpretability on commonsense reasoning research and other NLP tasks and on combining explanation with language generation. Our dataset is released at https://github.com/wangcunxiang/Sen-Making-and-Explanation.

\section{Related Work}
There has been datasets which focus on non-linguistic world knowledge validation \citep{wang2018modeling} or only limited attributes or actions of world knowledge \cite{verbphysics}

A widely used and important task is the Winograd Schema Challenge (WSC) \citep{WSC2012,WSC2015} that needs more commonsense knowledge. For example, \emph{``The trophy would not fit in the brown suitcase because it was too big (small). What was too big (small)?"
``Answer 0: the trophy",``Answer 1: the suitcase".}
However, WSC estimates common sense indirectly and it does not consider explanation on why one option is true while the other is wrong. 

Choice of Plausible Alternatives (COPA) \citep{COPA} puts emphasis on events and consequences. Each question of COPA asks to find the suitable cause or result of the premise from two given alternatives. All premises and alternatives are simple sentences. For example, \emph{``Premise: The man broke his toe. What was the CAUSE of this?"
``1: He got a hole in his sock.",
``2: He dropped a hammer on his foot.".}
Several subsequent datasets are inspired by COPA. The JHU Ordinal Common-sense Inference (JOCI) \citep{JOCI} aims to label the plausibility from 5 (very likely) to 1 (impossible) of human response after a certain situation. Situations with Adversarial Generations (SWAG) \citep{SWAG} requests to choose the most likely-to-happen alternative after a specific situation. 
Those datasets put emphasis on the pre-situations and/or the after-situations of certain situations, but not on the reasons why they occur or lead. 
 

Some datasets are inspired by reading comprehension, providing some textual materials and questions, asking to find suitable answers from the provided materials. The Story Cloze Test and ROCStories Corpora \citep{RocStories2016,RocStories2018} require to figure out the right ending from two candidate sentences after a four-sentence story. For a narrative text, MCScript \citep{MCScript} give various types of questions and pairs of answer candidates for each question. Most questions require knowledge beyond the facts mentioned in the text. Compared to those reading comprehension tasks, our benchmark encourages people to use any external resources they want as long as they will help in the task.

Some other datasets are evolved from QA problems and care more about factual commonsense knowledge. SQUABU \citep{SQUABU} provides a small hand-constructed test of commonsense and scientific questions. CommonsenseQA \citep{COMMONSENSEQA} asks crowd workers to create questions from ConceptNet \citep{ConceptNet5.5}, which is a large knowledge graph of commonsense knowledge, where each question discriminates its answer candidates between three target concepts that all share the same relationship to a single source drawn from ConceptNet
OpenBookQA \citep{OpenBookQA} provides questions and answer candidates, as well as thousands of diverse facts about elementary level science that are related to the questions. The AI2 Reasoning Challenge (ARC) \citep{ARC} gives thousands of questions with different knowledge types,
as well as a relevant 14M-sentence corpus, mixed with science facts and other narrative sentences.
Those questions are not easy to answer without specializing certain knowledge while our questions are easy for both adults and children. 

In contrast, to our work, the tasks above do not directly estimate common sense or ask the logical reasons behind the correct answers and questions.

In recent years, some large-scale commonsense inference knowledge resources have been released, which may be helpful in commonsense reasoning tasks.
Atomic \citep{ATOMIC} presents a huge everyday commonsense reasoning knowledge graph, which has nine \emph{if-then} relations with variables, including causes, effects, and so on. 
Event2Mind \citep{Event2Mind} proposes a new corpus and task, aiming to find out mentioned/unmentioned people's intents and reactions under various daily circumstances. These datasets are not directly useful for our benchmark since they focus only on a small domain. ConceptNet is a prestigious knowledge graph that has been upgraded over a long time \citep{ConceptNet,ConceptNet3,ConceptNet5,ConceptNet5.5}. ConceptNet constructs triples using labeled edges as relations and various words and/or phrases as entities. It also has the sentences describing the corresponding triples. Thus we consider using ConceptNet knowledge for fine-tuning our models.
 
\section{Testset}
\textbf{Task.}
Formally, each instance in our dataset is composed of 5 sentences: \{$s_1$, $s_2$, $r_1$, $r_2$, $r_3$\}. $s_1$ and $s_2$ are two similar sentences which in the same syntactic structure and differ by only few words, but only one of them makes sense while the other does not. They are used on our first subtask called \emph{Sen-Making}, which requires the model to identify which one is valid. For the invalid sentence, we have three optional reasons $r_1$, $r_2$ and $r_3$ to explain why the sentence is invalid. Our subtask 2, named \emph{Explanation}, requires that the only one correct reason be identified from two other confusing ones. We use the accuracy score to evaluate both subtasks.

\textbf{Data.}
For version 1.0 of our dataset, we have created 2,021 samples. Task 1 has 2,021 against-common-sense sentences, 2,021 correct sentences; Task 2 has 2,021 true reasons and 4,042 confusing reasons. We plan to release a dataset with more tasks and samples in the future.

\begin{figure*}[t]
  \center{\includegraphics[width=16cm]  {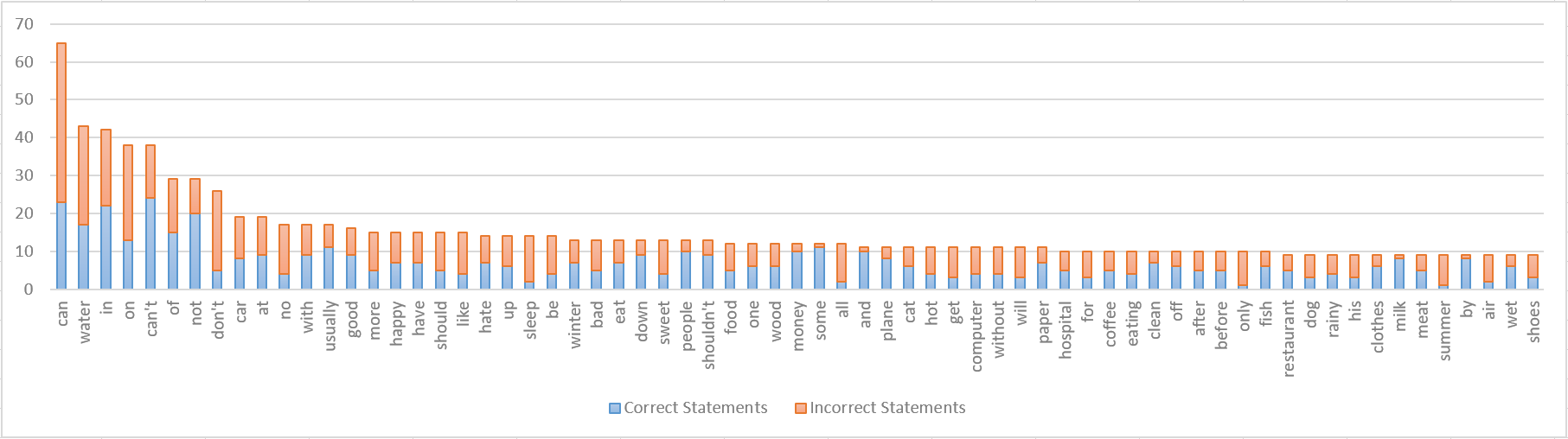}}  
  \caption{\label{1}  Number of `Different Words'}
  \vspace{-3.3mm}
\end{figure*}
\textbf{Annotation guidelines.}
When writing samples, annotators were asked to follow several principles. 
First, try to avoid complex knowledge and focus on daily common sense, and should make the questions as understandable as possible. Every literate person is able to give the right answers.
Second, the confusing reasons should better contain more important words like entities and activities in the against-common-sense statements, for example, the confusing reasons of  ``\emph{he put an elephant into the fridge}" should better contain both ``\emph{elephant}" and ``\emph{fridge}".
Third, we want the confusing reasons to be related to the statements and correct reasons and not deviate from the problem context, otherwise, it may be easily captured by BERT \citep{COMMONSENSEQA}, which models the sentence contexts explicitly. 
Next, the three option reasons should be only related to the incorrect sentence rather the correct sentence. Because we want further studies to be able to estimate against-common-sense statements without those correct statements. 
Furthermore, the confusing sentences should be correct themselves. Otherwise, the models may simply ignore the incorrect options without considering the casual relations between them. This worry was raised from the fact that models can achieve high performance in the ROC Story Cloze Task when only looking at the alternative endings ignoring the stories \citep{Schwartz2017}.
Last, we control the length of each sentence, making the incorrect statement nearly as long as the correct statement, and the right reason neither too long nor too short among the three reasons.
However, those principles are all soft restrictions; some samples are still good ones even without obeying those principles.  

\textbf{Annotation process.}
We ask data annotators to write samples by themselves. Then other researchers (the first and second author of this paper) will examine those samples case by case. We also provide them with external sources to stimulate inspiration, such as the raw English sentences of ConceptNet5.5 \citep{ConceptNet5.5}, but those inspirational sentences should not be used directly. For example, \emph{``he was sent to a (restaurant)/(hospital) for treatment after a car crash"} were inspired by the two sentences \emph{``restaurants provide food"} and \emph{``hospitals provide medical care"}
However, those corpus may have incorrect and one-sided knowledge, we do not use those sentences. Besides, we also let them get inspiration from existing commonsense reasoning questions like WSC \citep{WSC2012,WSC2015}, COPA \citep{COPA} and SQUABU \citep{SQUABU}. 

\textbf{Cautions when using the data.}
1. Researchers are encouraged to use what they deemed appropriate for the tasks to train their model, and use our testset for evaluation. 2. Please do not use the three optional reasons when performing \emph{Sen-Making} task, or the task will be artificially easy.

\subsection{Corpus Analysis}
The average length of two statements in the Sen-Making task are both 8.26, exactly the same.
The average length of true reasons is 7.63, slightly smaller than the confusing reasons' average length, which is 7.77.

We analyse the different words between correct statements and incorrect statements, like in ``He put (an elephant)/(a turkey) into the fridge'',  `an elephant' is `different word' in incorrect statements and `a turkey' is `different word' in correct statements. 
From `different words', we can get an overview about how correct statements and incorrect statements differ from each other.

We select the words which appear more than 9 times in the `different words' and remove stopwords like `a',`an' and `the' to draw Figure 2.
blue lines are the words appear in correct statements but do not appear in incorrect statements, while orange lines are opposite.
We find that incorrect statements have much the same negative different words compared with correct statements. The incorrect statements have 53 `don't' or `can't' or `not' or `no' as different words, while the correct statements have 55. That can illustrate that the corpus do not use negative words to construct incorrect statements or correct statements.

\section{Experiments}

\begin{table*}[t]
  \centering
  \label{label:results}
\begin{tabular}{|c|c|c|}
\hline
Model &  Sen-Making  &  Explanation  \\
\hline
Random &50.0\%&33.3\%\\
\hline
ELMo &69.4\% &33.4\%\\
\hline
BERT &70.1\%&45.6\%\\
\hline
fine-tuned ELMo &74.1\%&34.6\%\\
\hline
Human Performance &99.1\%&97.8\%\\
\hline
\end{tabular}

\caption{Experimental Results}
\vspace{-3mm}
\end{table*}

We choose state-of-the-art language models trained over large texts as our baselines, assuming that common sense knowledge are encoded over texts. For the sense making task, we calculate score of both statements, which is defined as 
\begin{equation}
  \begin{aligned}
  score = (p_{w_1}*p_{w_2}*...*p_{w_n})^ {(-1/n)}=\\
  (\prod_{i=1}^{n}P(w_{i} \mid w_{1}...w_{i-1}w_{i+1}...w_{n}))^{-1/n}
  \end{aligned}
\end{equation}
, then choosing the one with lower scores as the correct one. For explanation,  we first concatenate the statement with the each reason and then use the three concatenated sentences to calculate scores. For example, we concatenate \emph{``he put an elephant into the fridge"} with its optional reasons to be \emph{``he put an elephant into the fridge" is against common sense because an elephant cannot eat a fridge"}.

We also conduct human evaluation on our dataset. Each sample is answered by at least three testees. If more than half testees do one sample wrong (either \emph{Sen-Making} or \emph{Explanation}), we will rewrite or abolish the sample. Otherwise, we will keep it and record it in results.

\subsection{Results \& Analysis}

As shown in Table 1, ELMo and BERT have a significant advantage over random results in \emph{Sen-Making}. We conjure that both of them have the ability to judge whether a sentence is with or against common sense. 
For \emph{Sen-Making}, ELMo does better than BERT; However, BERT beats ELMo in \emph{Explanation}.
ELMo does poorly in \emph{Explanation}, much the same as a random guess, which shows that ELMo cannot handle the casual relationship between the incorrect statements and the reasons. 
In contrast, BERT is significantly better than random guess in \emph{Explanation}. This can be attributed to BERT's training on \emph{Next Sentence Prediction} \citep{BERT}, which can assist to handle the logic relationship between two sentences.  

Fine-tuned ELMo has an obvious improvement in \emph{Sen-Making} and a non-obvious improvement in \emph{Explanation}, probably because introducing knowledge will help models to identify common sense but cannot help them in inference. However, fine-tuning makes BERT perform the same in \emph{Sen-Making} and even worse in \emph{Explanation}. It is likely because the original BERT models trained on BookCorpus \citep{BookCorpus} and English Wikipedia contain sufficient common knowledge and the fine-tune operation may be useless or even makes the models be specific in the fine-tuning corpora; Besides, fine-tuning may corrupt the structure formed by \emph{Next Sentence Prediction}.  
How to overcome the weaknesses of fine-tuning operation may be an interesting topic in representation learning.

The method of fine-tuning can be viewed as a way to make use of human labelled knowledge base for common sense, however, a thorough investigation along this direction takes significant research effort, which is beyond the scope of this paper. It is also important to note that human labelled commonsense knowledge bases are inevitably limited by the scope. Therefore, gleaning knowledge from large raw text is still a more promising direction.

Note also that unlike reading comprehension tasks \cite{SQuAD}, where machines can surpass human performance by careful fine-tuning, it remains a big challenge for systems to reach human performance, which is near 100\% for sense making. When the human testee is asked to look at the mistakes that they make, we find that most human errors are due to reduced concentration rather than conceptual issues.

\subsection{Case Study}

We find an interesting case illustrating that fine-tuning operation can help model identify common sense but cannot help much in inference. The example is \emph{``New York is located in the northeastern part of USA"} vs \emph{``the USA is located in the northeastern part of New York"}.
The original ELMo is incorrect in both \emph{Sen-Making} and \emph{Explanation}. After fine-tuning with a corpora which contains that ``New York is a city" and ``the USA is a country", the ELMo can figure out the incorrect sentence but still cannot pick out the correct reason. 

We find that LM-based techniques cannot handle commonsense Sen-Making problems which need inference, such as the case \emph{``I'm too full for lunch, so I'm going (for a walk)/(to eat more)."} and its explanation, no matter fine-tuned or not.
This is likely because it remains difficult for LSTM language models for making multi-step inference.

\section{Conclusion}



We created a benchmark for directly evaluating whether a system has the capability of sense making and explanation, evaluating models trained over the large raw text as well as a common sense database on the test set. Results show that sense making remains a technical challenge for such models, whereas inference is a key factor that is missing.


\section*{Acknowledgement}
We thank the anonymous reviewers for constructive suggestions, and the non-author data annotators Run'ge Yan, Chenyan Du, Zinqun Zhou and Qikui Feng. The work is supported by NSFC grant number 61572245. Yue Zhang is the corresponding author.

\bibliography{commonsense}
\bibliographystyle{acl_natbib}

\end{document}